%% file: main.tex
\documentclass[11pt,letterpaper]{article}

\usepackage{cogsys}
\usepackage{times}
\usepackage{helvet}
\usepackage{courier}
 \usepackage{url} 
\usepackage{mdframed} 
\usepackage{amsmath} 
\usepackage{amssymb} 
\usepackage{color} 
\usepackage{proof}  
\usepackage{bussproofs}
\usepackage{graphicx}
\usepackage{subcaption}

\usepackage{paralist}

\usepackage[usenames,dvipsnames,svgnames,table]{xcolor}

\input{commands.tex}

\defsort{\Block}{Block}
\defsort{\Surface}{Surface}
\defsort{\Goal}{Goal}

\defsymbol{\on}{on}
\defsymbol{\clear}{clear}
\defsymbol{\goal}{goal}
\defsymbol{\stack}{stack}
\defsymbol{\unstack}{unstack}
\defsymbol{\inRoom}{inRoom}

\defconstant{\ctable}{$\mathit{table}$}
\defconstant{\ablock}{$A$}
\defconstant{\bblock}{$B$}
\defconstant{\cblock}{$C$}
\defconstant{\humana}{$h_1$}
\defconstant{\humanb}{$h_2$}
\defconstant{\cir}{$\mathit{cais}$}

\usepackage{natbib}
\cogsysheading{X}{20XX}{1-6}{X/20XX}{X/20XX}

\ShortHeadings{Toward Cognitive and Immersive Systems}{M.\ Peveler,
  N.\ S.\ Govindarajulu, S.\ Bringsjord, A.\ Sen, B.\ Srivastava, \& K.\ Talamadupula,}
              
\usepackage{lineno}

\usepackage{csquotes}

\newcounter{itemcounter}

\usepackage{tcolorbox}
\tcbuselibrary{skins}

\usepackage{tikz}
\usepackage{pagecolor}
\usetikzlibrary{decorations, decorations.text,backgrounds,fit}

\usetikzlibrary{positioning}
\makeatletter
\tikzset{
  auto centering/.style={execute at end picture={
      \node[fit=(current bounding box),minimum width=#1-2*\tikz@framexsep,inner sep=3,
      ]{};
    }},
  auto centering/.default=0.6\columnwidth,
}
\makeatother

\tikzset{block/.style={minimum size=0.65cm,outer sep=0pt,draw,rectangle,node distance=0pt}}
\tikzset{empty/.style={minimum size=1cm,outer sep=0pt,node distance=0pt}}

\begin{document}

\title{Toward Cognitive and Immersive Systems:\\
       Experiments in a Cognitive Microworld}

\author{Matthew Peveler}{pevelm@rpi.edu}
\author{Naveen Sundar Govindarajulu}{Naveen.Sundar.G@gmail.com}
\author{Selmer Bringsjord}{Selmer.Bringsjord@gmail.com}
\author{Atriya Sen}{atriya@atriyasen.com}
\address{Rensselaer AI \& Reasoning Lab, Department of Computer Science, \\
Rensselaer Polytechnic Institute (RPI), Troy, NY 12180, USA}
\author{Biplav Srivastava}{biplavs@us.ibm.com}
\author{Kartik Talamadupula}{krtalamad@us.ibm.com}
\author{Hui Su}{huisuibmres@us.ibm.com}
\address{IBM TJ Watson Research Center, Yorktown Heights, NY 10598, USA}
\vskip 0.2in

\input{abstract}






\newpage
\input{introduction}

\input{cognitive_event_calculus}

\input{requirements}


\input{satisfied}

\input{cognitive_blockworld_framework}

\input{scenario}

\input{objections}

\input{conclusion}

\input{acknowledgments}

{\parindent -10pt\leftskip 10pt\noindent
\bibliographystyle{cogsysapa}
\bibliography{bibliography,naveen,main72}
}

\end{document}

%% file: commands.tex
\newcommand{\lsort}[1]{%
  \ensuremath{\mbox{\textsf{#1}}}}
\newcommand{\defsort}[2]{%
  \newcommand{#1}{\lsort{#2}}}
\defsort{\Action}{Action}
\defsort{\Time}{Time}
\defsort{\Self}{Self}

\defsort{\Agent}{Agent}
\defsort{\Entrant}{Entrant}
\defsort{\ActionType}{ActionType}
\defsort{\Moment}{Moment}
\defsort{\Boolean}{Formula}
\defsort{\PayOut}{PayOut}
\defsort{\Fluent}{Fluent}
\defsort{\Event}{Event}
\defsort{\Object}{Object}
\defsort{\RealTerm}{RealTerm}

\defsort{\Numeric}{Numeric}
\defsort{\Number}{Number}
\defsort{\Trolley}{Trolley}
\defsort{\Track}{Track}
\defsort{\Moveable}{Moveable}

\newcommand{\lsymbol}[1]{%
  \ensuremath{\mathit{#1}}}
\newcommand{\defsymbol}[2]{%
  \newcommand{#1}{\lsymbol{#2}}}
\defsymbol{\action}{action}
\defsymbol{\initially}{initially}
\defsymbol{\holds}{holds}
\defsymbol{\happens}{happens}
\defsymbol{\clipped}{clipped}
\defsymbol{\initiates}{initiates}
\defsymbol{\terminates}{terminates}
\defsymbol{\prior}{prior}
\defsymbol{\interval}{interval}

\defsymbol{\does}{does}
\defsymbol{\plans}{plans}
\defsymbol{\act}{act}
\defsymbol{\react}{react}
\defsymbol{\payTot}{pay_{tot}}
\defsymbol{\fight}{fight}
\defsymbol{\coop}{coop}
\defsymbol{\enter}{enter}
\defsymbol{\stayout}{stayout}
\defsymbol{\learns}{learns}
\defsymbol{\payoff}{payoff}
\defsymbol{\position}{position}
\defsymbol{\dead}{dead}
\defsymbol{\damaged}{damaged}

\defsymbol{\onrails}{onrails}
\defsymbol{\switch}{switch}
\defsymbol{\drop}{drop}
\defsymbol{\vicinity}{vicinity}

\newcommand{\lconstant}[1]{%
  \ensuremath{\mbox{\textsf{#1}}}}
\newcommand{\defconstant}[2]{%
  \newcommand{#1}{\lconstant{#2}}}
\defconstant{\Enter}{Enter}
\defconstant{\StayOut}{StayOut}
\defconstant{\Fight}{Fight}
\defconstant{\Acquiesce}{Acquiesce}
\defconstant{\cs}{cs }

\newcommand{\lmodality}[1]{%
  \ensuremath{\mathbf{#1}}}
\newcommand{\defmodality}[2]{%
  \newcommand{#1}{\lmodality{#2}}}
\defmodality{\common}{C}
\defmodality{\knows}{K}
\defmodality{\believes}{B}
\defmodality{\perceives}{P}
\defmodality{\mental}{M}
\defmodality{\requests}{R}

\defmodality{\desires}{D}
\defmodality{\intends}{I}
\defmodality{\says}{S}
\defmodality{\ought}{O}

\newcommand{\DCEC}{\ensuremath{\mathcal{DCEC}}}

\newcommand{\lif}{\rightarrow}

\newcommand{\sep}{\ \lvert \ }

\usepackage{booktabs}
\usepackage{mdframed}
\usepackage{paralist}

\usepackage{color}

\usepackage[usenames,dvipsnames,svgnames,table]{xcolor}

\newcommand{\type}[1]{\textbf{#1}}

%% file: abstract.tex
\begin{abstract}
  As computational power has continued to increase, and sensors have
  become more accurate, the corresponding advent of systems that are
  at once cognitive and immersive has arrived.  These
  \textit{cognitive and immersive systems} (CAISs) fall squarely into
  the intersection of AI with HCI/HRI: such systems interact with and
  assist the human agents that enter them, in no small part because
  such systems are infused with AI able to understand and reason about
  these humans and their knowledge, beliefs, goals, communications,
  plans, etc.  We herein explain our approach to engineering CAISs.
  We emphasize the capacity of a CAIS to develop and reason over a
  ``theory of the mind'' of its human partners.  This capacity entails
  that the AI in question has a sophisticated model of the beliefs,
  knowledge, goals, desires, emotions, etc.\ of these humans.  To
  accomplish this engineering, a formal framework of very high
  expressivity is needed.  In our case, this framework is a
  \textit{cognitive event calculus}, a particular kind of quantified
  multi-operator modal logic, and a matching high-expressivity
  automated reasoner and planner.  To explain, advance, and to a
  degree validate our approach, we show that a calculus of this type
  satisfies a set of formal requirements, and can enable a CAIS to
  understand a psychologically tricky scenario couched in what we call
  the \textit{cognitive polysolid framework} (CPF).  We also formally
  show that a room that satisfies these requirements can have a useful
  property we term \emph{expectation of usefulness}.  CPF, a sub-class
  of \textit{cognitive microworlds}, includes machinery able to
  represent and plan over not merely blocks and actions (such as seen
  in the primitive ``blocks worlds'' of old), but also over agents and
  their mental attitudes about both other agents and inanimate
  objects.
\end{abstract}


%% file: introduction.tex
\section{Introduction}
\label{sect:intro}
\vspace{2mm}

In contemporary AI research devoted to decision support, the challenge
is often taken to be that of providing AI support to a single human.
However, much human problem-solving
is fundamentally social, in that a group of people must work together
to solve a problem, and must rely upon machine intelligence that is
itself highly diverse.  Examples of such activities include: hiring a
person into a university or company, tackling an emergency crisis like
a water pipeline break, planning an intricate medical operation,
deciding on companies for merging or acquisition, etc.  Motivated by
such challenges, we are interested in how an artificial agent ---
embedded in a social-collaboration environment like an immersive room
--- can, on the spot, help a group of human participants.

\begin{figure}
\centering
\includegraphics[width=0.5\columnwidth]{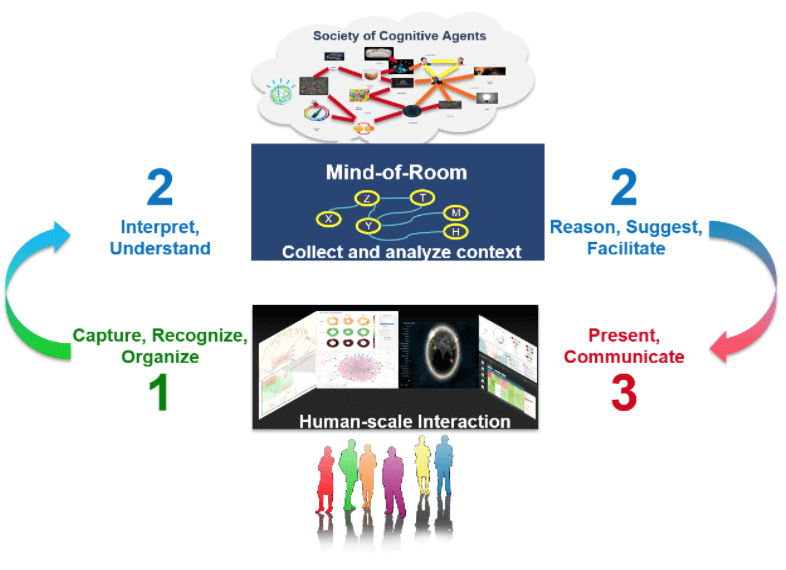}
\caption{The flow of information through the areas of a CAIS.}
\label{fig:cycle-cais}
\end{figure}

We first introduce the notion of a \emph{cognitive and immersive
  system} (CAIS), which is comprised of three sub-areas linked to each
other in a cyclical flow, shown in Figure \ref{fig:cycle-cais}.  The
first area is responsible for perception and sensing within the
environment housing the human agents (specifically for us, within the
room).  Percepts come courtesy of a range of sensors; for example,
microphones and kinects.  The second area covers interpreting,
understanding, and acting upon perceived data through reasoning,
planning, learning, and NLP.  The third area is devoted to displaying
both percepts, and the results of processing thereof, in rich
multi-modal ways.  The particular CAIS we have so far used for our
investigations
%
additionally has access to a variety of external machines and services
that can be called upon to process requests, queries, and tasks, and
to incorporate the results of analysis of additional information from
these ines into further decision-making.  An important part of our
particular CAIS is that there are some number of overseeing AIs
(agents) operating at the system level that can make use of any part
of the CAIS to assist and aid the humans and other AIs that are
operating within the room.  Thus this architecture is neither fully
centralized, nor fully distributed, but aims to combine the strengths
of both.

As an initial test of our CAIS implementation, we examine two
scenarios wherein participants have an imbalance of knowledge/beliefs
that might influence their actions.  It is common in group discussions
that not all participants are aware of everything said in the
discussion, whether because they missed something being said or
misconstrue something.  However, they may still try to act on their
beliefs that do not (or no longer) properly reflect reality, or may
know of their ignorance and wish to remedy it.  A CAIS should be able
to offer help to participants in these cases, either by alerting
participants if they are acting on beliefs that are no longer
relevant, or by giving a brief summary of the things that had happened
while they were out of the discussion.  To do this, a CAIS must be
able to model the theory of mind
\citep{PremackWoodruff78,frith2005,ka_sb_scc_seqcalc} of the
participants and track its state through time.  On the basis of its
understanding of this modeling, a CAIS must
be able to step in as necessary to offer corrective advice, along with
a justification for it.  Most importantly, the particular capacities
we have just enumerated as desiderata for a CAIS must flow from
underlying formal requirements that rigorously capture the general
desiderata in question.

To accomplish the above tasks, we first present informal requirements
that differentiate a CAIS from other intelligent agents. The
requirements are in terms of the cognitive states and common knowledge
of the agents within a CAIS. As far as we know, this is the first such
characterization of what separates an intelligent room from an
intelligent agent.  We then cast the informal requirements in a formal
logic and show that these requirements lead to a useful property
(\textbf{Property 1}).  We then briefly present a framework for a
domain of problems that can be used to test a CAIS.  From there, we
define two tasks with that framework, and then show that by
\textbf{Property 1}, the system (with the relevant information) can
solve the tasks.  Finally, after a sustained discussion of related,
prior research in AI in which we make clear the unique power of our
formalisms and technology, we discuss promising future lines of work.


\input{cognitive_immersive_room}

\input{related_work}


%% file: cognitive_immersive_room.tex
\subsection{Cognitive Immersive Room}

In development of a CAIS, much work has gone into the creation of a
cognitive immersive room architecture~\citep{divekar2018} which allows
for research into augmenting human group collaboration and
decision-making with cognitive artificial agents.  To start, we build
on the prior work in the space of intelligent rooms, primarily that
carried out in \cite{brooks1997intelligent}.

At the core of our room, we have an array of microphones that hear
what people say, which utilizes a transcription service to translate
the speech to text.  Given the text, we then check for the presence of
a ``trigger word'' to indicate if a user is talking to the room or
not.  If the ``trigger word'' is detected, the text is then further
analyzed (utilizing the IBM Watson
Assistant\footnote{https://www.ibm.com/watson/services/conversation/})
to extract an intent for the text as well as any keywords in the text
given the command.  The intent and keywords are then fed into the
executor which in turn drives the room, and can call external services
as necessary and output content to users via connected displays or
speakers.  Additionally, the executor maintains the state of the room
as well as definition of the conversation tree that is available to
the participants at any given time.  Each of these components are
implemented and run separately and communication between them is
handled via passing JSON objects using the message queue server
RabbitMQ\footnote{https://www.rabbitmq.com}.

In addition to these, there is a host of external web services made
available to any component through the standard GET/POST HTTP request
headers.  An example of such a web service is the
``name-resolver-general,'' which, given a name, will return a list of
probable matches for that keyword, which allows us to handle
misspellings introduced in the keyword by the speech-to-text
translation.  Each of these external web services are registered in
our ``service-registry,'' which the executor references using its name
to get the IP and port of these services to be able to call them.  The
registry also acts as a monitor for each service to determine if the
service is available or not.  Finally, we utilize the in-memory data
store of Redis\footnote{https://redis.io} to store information about
our agents during a particular usage or session of the room and then
use the PostgreSQL\footnote{https://www.postgresql.org} database to
store information on a longer-term basis.  The displays for the room
are three projectors on which we run
Electron\footnote{https://electronjs.org}, which allows for building
desktop GUIs powered by web technologies, including HTML, CSS, and JS.
In Electron, the executor can open various web pages and sites within
the GUI, showing both internal (such as transcript or command log) and
external content (such as Google maps or Youtube).

%% file: related_work.tex
\subsection{Recognition of the Need for Theory-of-Mind-Level AI}
\label{subsect:recognition_need}

Independent of the specific formalisms and corresponding technology
that we soon bring to bear herein, we first point out that the need to
model the mental states of humans in order to engineer certain AI
systems that understand and interact well with these humans has been
recognized.  For example, work in human-robot teaming has focused on
the use of automated planning techniques that take human goals and
(mental) states into account.  In addition, work on human-aware task
planning for mobile robots \citep{cirillo2009human} has used
\emph{predicted} plans for the humans to guide the automated system's
own planning.  This direction was made more explicit in work on
co\"{o}rdinating the goals and plans of humans and robots
\citep{talamadupula2014coordination,chakraborti2015planning}; here, a
subset of the humans' mental states relevant to the autonomous
system's planning problems was explicitly represented and reasoned
with.  Very recent work has focused on adapting these previous ideas
and techniques to proactive decision-making
\citep{radar2017aaaifss,kim2017aaaifss} and smart-room environments
\citep{jones2017aaaifss}.  \cite{pearce_etal_social_planning_aaai2014}
note the importance of what they dub ``social planning,'' which
includes an agent's seeking a goal via the modification of the mental
states of others.  For a final example, in \citep{langley2016pug},
PUG, a system that uses mental simulation to plan for symbolic goals
that have numeric utilities, is presented.\footnote{Our latter two
  examples here are briefly returned to later in ``Prior/Related Work
  and Novelty.''}

On the other hand, while these papers confirm the recognition to which
we refer, and while they feature some level of formalization of
human-level mental states, they seem to us to lack the necessary
technical formal and computational machinery needed to mechanize a
full human-level theory of mind --- let alone such a theory of mind
\emph{and} the requirements of a truly smart room/CAIS we set
below.\footnote{Note that in the present paper there is a limit to the
  theory-of-mind-modeling ``power'' we insist an overseeing AI have.
  E.g., we don't require that an AI overseeing an environment
  populated with humans have so-called \textit{phenomenal
    consciousness}, a form of ``what's it's like to'' consciousness
  characterized e.g.\ by \cite{bbs.block}, and claimed by
  \cite{sb_billion_conscious_robot} to be impossible for a mere
  machine to possess.  In sharp contrast with phenomenal
  consciousness, \textit{cognitive consciousness} consists only in the
  logico-mathematical \emph{structure} of human-level (and, indeed,
  above) cognition, instantiated through time.  While cognitive
  consciousness can be characterized axiomatically with help from the
  formal languages we introduce below for cognitive calculi
  \citep{axiomatizing_consciousness1}, in the present paper we do not
  require the AI overseeing CAISs to have even cognitive
  consciousness, and we specifically do not require, at this early
  point in our work on CAISs, cognitive \emph{self}-consciousness,
  despite the fact that the latter is something that has been
  significantly mechanized and implemented
  \cite{roman2015_robot_self-con,sb_on_knowledge_game}.}  We now turn
to the presentation of the requisite formal and computational
machinery.
%


%% file: cognitive_event_calculus.tex
\section{The Deontic Cognitive Event Calculus}

To capture the room in a formal way, we employ the \textbf{deontic
  cognitive event calculus} (\DCEC).\footnote{We do not use the
  deontic components in \DCEC\ in this paper.}  While the full syntax
and inference schemata are outside the scope of this paper, we give a
brief overview.\footnote{For a more in-depth primer on the \DCEC, see
  the appendix in \cite{nsg_sb_dde_ijcai}.} \DCEC\ is a multi-sorted
quantified modal logic with a well-defined syntax and proof calculus.
\DCEC\ subsumes the event calculus~\citep{mueller2014}, a first-order
calculus used for modeling events and actions and their effects upon
the world.  The proof calculus of \DCEC\ is based on natural deduction
\citep{gentzen1964} and includes all the introduction and elimination
rules for first-order logic, as well as inference schemata for the
modal operators and related structures.  \DCEC\ is a sorted system and
includes the following built-in sorts:

\begin{scriptsize}
\begin{center}
\begin{tabular}{lp{5.8cm}}  
\toprule
Sort    & Description \\
\midrule
\type{Agent} & Human and non-human actors.  \\

\type{Time} &  Time points. E.g., $t_i$, $birthday(son(jack))$ etc. \\

  \type{Event} & Used for events in the domain. \\
  \type{ActionType} & Abstract actions
                      instantiated by agents.\\
  \type{Action} & Events that occur
                  as actions by agents \\
  \type{Fluent} & Representing states of the world.\\
  \bottomrule
\end{tabular}
\end{center}
\end{scriptsize}

The intensional operators and necessary inference schemata for this
paper are shown below below.  The operator $\believes(a, t, \phi)$
represents that agent $a$ at time $t$ believes $\phi$.  The operator
$\knows(a, t, \phi)$ represents that agent $a$ at time $t$ knows
$\phi$.\footnote{Note that knowing is not decomposable or inferable
  from belief.}  The operator $\desires(a, t, \phi)$ represents that
agent $a$ at time $t$ desired $\phi$.  The operator $\common(t, \phi)$
represents that at time $t$, $\phi$ is common knowledge, which from
the inference schemata defined below we see means subsequently that
all agents know $\phi$.  The operator $\says(a, b, t, \phi)$
represents that agent $a$ told agent $b$ $\phi$ at time $t$.
Alternatively, it can be used as $\says(a, t, \phi)$, which represents
that agent $a$ at time $t$ said $\phi$ (and everyone hears it).  We
also have the operator $\perceives(a, t, \phi)$, which represents that
agent $a$ at time $t$ perceived $\phi$ (giving us also that agent $a$
knows $\phi$ at time $t$).

For our current purposes, the main inference schemata needed include
$I_{\knows}$ and $I_{\believes}$, which state that knowledge and
belief are closed under the inference system of $\DCEC$.  We also have
inference schemata that let us go from perception to knowledge
($I_1$), knowledge to belief ($I_2$), common knowledge to knowledge
($I_3$), and from knowledge to propositions that hold ($I_4$). Later
below, we also use \emph{derived inference schemata} for converting
perceptions to knowledge, knowledge to belief, common knowledge to
belief etc.,~labeled as $D_{[\perceives \leadsto \knows]}$,
$D_{[\knows \leadsto \believes]}$, and
$D_{[\common \leadsto \believes]}$ respectively
\citep{ArkoudasAndBringsjord2008Pricai}.

  \begin{scriptsize}
\begin{equation*}
 \begin{aligned}
&{\mbox{\textbf{Syntax}}}\\
    \mathit{S} &::= 
    \begin{aligned}
      & \Object \sep \Agent \sep \Self \sqsubset \Agent \sep
      \ActionType \\ & \Action \sqsubseteq
      \Event \sep \Moment \sep \Boolean \sep \Fluent \sep \Numeric\\
    \end{aligned}  
    \\ 
    \mathit{f} &::= 
    \begin{aligned}
      & \action: \Agent \times \ActionType \rightarrow \Action \\
      &  \initially: \Fluent \rightarrow \Boolean\\
      &  \holds: \Fluent \times \Moment \rightarrow \Boolean \\
      & \happens: \Event \times \Moment \rightarrow \Boolean \\
      & \clipped: \Moment \times \Fluent \times \Moment \rightarrow \Boolean \\
      & \initiates: \Event \times \Fluent \times \Moment \rightarrow \Boolean\\
      & \terminates: \Event \times \Fluent \times \Moment \rightarrow \Boolean \\
      & \prior: \Moment \times \Moment \rightarrow \Boolean\\
    \end{aligned}\\
        \mathit{t} &::=
    \begin{aligned}
      \mathit{x : S} \sep \mathit{c : S} \sep f(t_1,\ldots,t_n)
    \end{aligned}
    \\ 
    \mathit{\phi}&::= 
    \begin{aligned}
     & t:\Boolean \sep  \neg \phi \sep \phi \land \psi \sep \phi \lor
     \psi \sep \\
     & \perceives (a,t,\phi)  \sep \knows(a,t,\phi) \sep
     \common(t,\phi) \sep \says(a,b,t,\phi) \sep \\
     &  \says(a,t,\phi)  \sep \believes(a,t,\phi) \sep \desires(a,t,\holds(f,t')) \sep
     \intends(a,t,\phi)      \end{aligned}
  \end{aligned}
\hspace{20pt}\begin{aligned}
&{\mbox{\textbf{Inference Schemata} (fragment)}}\\
  &\infer[{[I_{\knows}]}]{\knows(a,t_2,\phi)}{\knows(a,t_1,\Gamma), \ 
    \ \Gamma\vdash\phi, \ \ t_1 \leq t_2} \\
& \infer[{[I_{\believes}]}]{\believes(a,t_2,\phi)}{\believes(a,t_1,\Gamma), \ 
    \ \Gamma\vdash\phi, \ \ t_1 \leq t_2} \\
 &\infer[{[I_1]}]{\common(t,\perceives(a,t,\phi)
   \lif\knows(a,t,\phi))}{}\\
&  \infer[{[I_2]}]{\common(t,\knows(a,t,\phi)
    \lif\believes(a,t,\phi))}{}\\
  &\infer[{[I_3]}]{\knows(a_1, t_1, \ldots
    \knows(a_n,t_n,\phi)\ldots)}{\common(t,\phi) \ t\leq t_1 \ldots t\leq
    t_n}\\
&  \infer[{[I_4]}]{\phi}{\knows(a,t,\phi)}\\
&\hspace{40pt}\vdots \hspace{40pt}\vdots
\end{aligned}
\end{equation*}
\end{scriptsize}
%

\subsection{Non-modal Systems are not Enough}
\label{sect:prop_not_enough}
Note that first-order logic is an \textbf{extensional} system; modal
logics are \textbf{intensional} systems.  \DCEC\ is
\textbf{intensional} in the sense that it has intensional
operators.\footnote{Please note that there is a vast difference
  between intension and intention.}  Formal systems that are
intensional are crucial for modeling theory-of-mind reasoning.  One
simple reason is that using first-order logic leads to unsound
inferences as shown below, in which we have an agent $r$ that knows
the manager of a team is the most responsible person in the team.
Agent $r$ does not know that $\mathit{Moe}$ is the manager of the
team, but it's true that $\mathit{Moe}$ is the manager.  If the
knowledge operator $\mathbf{K}$ is a simple first-order predicate, we
get the proof shown below, which produces a contradiction (that $r$
knows that $\mathit{Moe}$ is the manager) from true premises.  This
unsoundness persists even with more robust representation schemes in
extensional logics \citep{selmer_naveen_metaphil_web_intelligence}.

\vspace{10pt}

\begin{minipage}[b]{0.7\textwidth}
\begin{scriptsize}
 \begin{scriptsize}
\begin{equation*}
\begin{aligned}
&\fbox{1}\ \ \mathbf{K}\left(r,\
  \mathsf{Manager}\left(\mathit{team}, \mathit{mostResponsible}\left(\mathit{team}\right) \right)\right) \mbox{
  {\color{gray}; given}} \\
&\fbox{2}\ \ \lnot \mathbf{K}\left(r,\mathsf{Manager}\left(\mathit{team}, \mathit{Moe}\right)\right) \mbox{
  {\color{gray}; given}}\\
&\fbox{3}\ \ \mathit{Moe} = \mathit{mostResponsible}\left(\mathit{team}\right)  \mbox{
  {\color{gray}; given}}\\
&\fbox{4}\ \ \mathbf{K}\left(r,\mathsf{Manager}\left(\mathit{team}, \mathit{Moe}\right)\right)  \mbox{
  {\color{gray}; first-order inference from \fbox{3} and \fbox{1}}}\\
& \fbox{5}\ \ \mathbf{\bot}  \mbox{
  {\color{gray}; first-order inference from \fbox{4} and \fbox{2}}}
\end{aligned}
\end{equation*}
\end{scriptsize}
\end{scriptsize}
\end{minipage}

\input{reasoner}

\input{planner}

%% file: reasoner.tex
\subsection{Reasoner (Theorem Prover)}

To handle reasoning within \DCEC\, we utilize a quantified modal logic
theorem prover, \textsf{ShadowProver}.\footnote{The prover is
  available in both Java and Common Lisp and can be obtained at:
  \url{https://github.com/naveensundarg/prover}. The underlying
  first-order prover is SNARK, available at:
  \url{http://www.ai.sri.com/~stickel/snark.html}.}  The prover works
by utilizing a technique called \textbf{shadowing} to achieve speed
without sacrificing consistency in the system.  Describing the details
of the reasoner are beyond the scope here.  See
\citep{nsg_sb_dde_ijcai,uncertaintyized_cognitive_calculus} for more
details.

%% file: planner.tex
\subsection{Planner}
\label{subsect:planner}

Planning for the room is handled by \textsf{Spectra}, a planner based
on an \emph{extension} of STRIPS-style
planning.\footnote{``STRIPS-style'' ascribed to a planner $P$ means
  that some of the prominent properties that a STRIPS planner has, $P$
  has.  For illustration by analogy, a theorem prover in the style of
  some famous first-order one (e.g.\ Otter) could be capable of
  reasoning with second-order formulae.  A programming language in the
  logic-programming style could actually fail to be based on
  first-order logic or a fragment thereof.  It is important to note
  that \textsf{Spectra}'s planning is based on an \emph{extension} of
  STRIPS-style planning.} In this planning formalism, arbitrary
formulae of \DCEC\ are allowed in states, actions, and goals.  For
instance, valid states and goals can include: \emph{``No three blocks
  on the table should be of the same color.''}  and \emph{``Jack
  believes that Jill believes there is one block on the table.''}

%% file: requirements.tex
\section{What is a CAIS? Informal Requirements}
\label{sect:reqs}
Note that a CAIS can be considered an intelligent room that
specifically requires intelligence of a cognitive sort; that is, it is
not sufficient that the room be intelligent about, for example, search
queries over a domain $D$; the room should also be intelligent about
cognitive states of agents in the room and their cognitive states and
attitudes toward $D$.

Despite there being a significant amount of work done in building
intelligent environments (of varying levels of intelligence;
\citealp{coen1998design, brooks1997intelligent,chan2008review}), there
is no formalization of what constitutes an intelligent room and what
separates it from an intelligent agent.  Though \citep{coen1998design}
briefly differentiates an intelligent room from ubiquitous computing
based on the non-ubiquity of sensors in the former, there is not any
formal or rigorous discussion of what separates an intelligent room
from a mobile robot that roams around the room with an array of
sensors.  We offer below a sketch of informal requirements that an
immersive room should aim for.  Then we instantiate these requirements
using \DCEC.

The requirements in question are cognitive in nature and exceed
intelligent rooms with sensors that can answer queries over simple
extensional data (e.g.\ a room that can answer financial queries such
as \textit{``Show me the number of companies with revenue over X?''}).
At a high-level, we require that the two conditions below hold:

\begin{footnotesize}
  \begin{mdframed}[frametitle= Informal Requirements,
    frametitlebackgroundcolor=gray!25, nobreak, linecolor=white,backgroundcolor=gray!10]
    \begin{enumerate}
    \item[$\mathcal{C}$ \emph{Cognitive} ] A CAIS system should be
      able to help agents with cognitive tasks and goals.  For
      instance, a system that simply aids in querying a domain $D$ is
      not cognitive in nature; a system that aids an agent in
      convincing another agent that some state-of-affairs holds in $D$
      is considered cognitive.
    \item[$\mathcal{I}$ \emph{Immersive/Non-localized} ] There should
      be some attribute or property of a CAIS that is non-localized
      and distinguished from agents in the room.  Moreover, this
      property should be \textbf{common knowledge}.  (Note: this is
      not easily achievable with a physical robot, and this condition
      differentiates a CAIS system from a cognitive
      agent.)\footnote{This condition may not strictly be realizable,
        but the goal is to at a minimum build systems that approach
        this ideal condition.}
    \end{enumerate}
  \end{mdframed}
\end{footnotesize}

\section{Formal Requirements for a CAIS}
\label{sect:freqs}
Now we translate the informal requirements presented above for a CAIS
$\gamma$.  The CAIS we present acts as an arbiter when goals of agents
conflict, and acts to rectify false or missing beliefs.  Assume that
enclosed within $\gamma$'s space at time $t$ are agents $A(t) = \{a_1,
\ldots, a_n\}$.  (Note: $\gamma$ is also an agent but is not included
in $A(t)$.)  Time is assumed to be discrete, as in the discrete event
calculus presented in \cite{mueller2014}.  There is a background set
of axioms and propositions $\Gamma(t)$ that is operational at time
$t$.  We have a fluent $\vicinity$ that tells us whether an agent is
in the vicinity of a fluent, event, or another agent.  Only events and
fluents in the vicinity of an agent can be observed by the agent.

\begin{footnotesize}
$$\vicinity: \Agent \times \Fluent \cup \Event \cup \Agent \rightarrow
\Fluent $$
\end{footnotesize}
For the cognitive condition $\mathcal{C}$ above, we have the following
concrete requirements that we implement in our system.  Later, we give
examples of these requirements in action.

  \begin{footnotesize}
    \begin{mdframed}[frametitle= Formal Requirements for $\mathcal{C}$ ,
      frametitlebackgroundcolor=gray!25, nobreak,
      linecolor=white,backgroundcolor=gray!10]
      Assume $\Gamma \vdash t < t + \Delta$
      \begin{enumerate}
      \item[$\mathbf{C}^f_1:$] It is common knowledge that, if an
        agent $x$ has a false belief, $\gamma$ informs the agent of
        the belief:\footnote{Please note that inference in \DCEC\ is
          non-monotonic as it includes the event calculus, which is
          non-monotonic.  If an agent $a$ believes $\phi$ based on
          prior information, adding new information can cause the
          agent to not believe $\phi$.}
         \begin{equation*}
          \common\color{gray!80}\left(\color{black}t, \left[\begin{aligned}
                &  \believes(\gamma, t, \phi) \land \believes(\gamma, t,
                \believes\big(x, t, \lnot \phi)\big) 
                \\ & \hspace{50pt} \rightarrow \\
                & \hspace{30pt}\says(\gamma, x, t + \Delta, \phi)
              \end{aligned}\right]\color{gray!80}\right)\color{black}
        \end{equation*} 

      \item[$\mathbf{C}^f_2:$] It is common knowledge that, if an
        agent $x$ has a missing belief, $\gamma$ informs the agent of
        the belief:
        \begin{equation*}
          \common\color{gray!80}\left(\color{black}t,\left[ \begin{aligned}
                &  \believes(\gamma, t, \phi) \land \believes(\gamma, t,
                \lnot \believes\big(x, t, \phi)\big) 
                \\ &  \hspace{50pt}  \rightarrow \\
                & \hspace{30pt}\says(\gamma, x, t + \Delta, \phi)
              \end{aligned}\right]\color{gray!80}\right)\color{black}
        \end{equation*} \end{enumerate}
    \end{mdframed}
  \end{footnotesize}

  For the immersive condition $\mathcal{I}$ above, we have the
  following conditions:

  \begin{footnotesize}
    \begin{mdframed}[frametitle= Formal Requirements for $\mathcal{I}$ ,
      frametitlebackgroundcolor=gray!25, nobreak,
      linecolor=white,backgroundcolor=gray!10]
      \begin{enumerate}
      \item[$\mathbf{I}^f_1:$] It is common knowledge that at any
        point in time $t$ an agent $x$, different from the CAIS system
        $\gamma$, can observe events or conditions (fluents) only in
        its vicinity.
        \begin{equation*}
          \begin{aligned}
            \common\Bigg(\forall x, t, f:& (x \not = \gamma) \rightarrow \Big[\perceives\big(x, t, \holds(f, t)\big) \rightarrow
            \holds(\vicinity(x, f), t)\Big]\Bigg)\\
            \common\Bigg(\forall x, t, e:& (x \not = \gamma) \rightarrow \Big[\perceives\big(x, t, \happens(e, t)\big) \rightarrow
            \holds(\vicinity(x, e), t)\Big]\Bigg)\\
          \end{aligned}
        \end{equation*} 

      \item[$\mathbf{I}^f_2:$] It is common knowledge that actions
        performed by the agent are in its vicinity.
        \begin{equation*}
          \common\Big(\forall x, \alpha:  \holds\big(\vicinity(x, \action(a, \alpha)), t\big)\Big)
        \end{equation*} 
      \item[$\mathbf{I}^f_3:$] It is common knowledge that all events
        and fluents are perceived by $\gamma$.  This is represented by
        the four conditions below:
        \begin{equation*}
          \begin{aligned}
       &  (i) \hspace{10pt}  \common\Bigg(\forall  t, f:  \Big[\holds(f, t) \leftrightarrow \perceives\big(\gamma, t, \holds(f, t)\big) \Big]\Bigg)\\
       & (ii)  \hspace{10pt}   \common\Bigg(\forall  t, e: \Big[\happens(e, t)
            \leftrightarrow \perceives\big(\gamma, t, \happens(e,
            t)\big) \Big]\Bigg)\\
        &   (iii)  \hspace{10pt}   \common\Bigg(\forall  t, f:  \Big[\lnot \holds(f, t)
            \leftrightarrow \perceives\big(\gamma, t, \lnot \holds(f, t)\big) \Big]\Bigg)\\
        &   (iv) \hspace{10pt}    \common\Bigg(\forall  t, e:\Big[\lnot \happens(e, t)
            \leftrightarrow \perceives\big(\gamma, t, \lnot \happens(e, t)\big) \Big]\Bigg)
          \end{aligned}
        \end{equation*} 
      \end{enumerate}
    \end{mdframed}
  \end{footnotesize}


%% file: satisfied.tex
\section{A Foundational Property of CAIS}
\label{sect:prop}
One of the benefits of having a properly designed CAIS is that humans
inside it can rely upon the CAIS to help other agents with relevant
missing or false information $\phi$, as opposed to the case with a
mobile robot.  If we had a localized mobile robot instead of a CAIS,
the human would need to decide whether the robot has the required
information $\phi$ and needs to believe that the robot believes that
the other human is missing the relevant information $\phi$.  If a CAIS
system satisfies the above condition, we can derive the following
foundationally important (object-level in \DCEC) property that states
this in a formal manner.

\begin{small}
\begin{mdframed}[linecolor=white, frametitle= Property 1: Expectation of Usefulness , frametitlebackgroundcolor=gray!25,
  backgroundcolor=gray!10, nobreak=true ,roundcorner=8pt]
  If the above properties hold, then an agent $a$ that perceives that
  another agent $b$ is not aware of an event happening, believes that
  CAIS $\gamma$ will inform $b$ (Assume: $\Gamma \vdash t < t + \Delta$):
\begin{equation*}
\begin{aligned}
&\perceives\big(a, t, \happens\big(e, t\big)\big) \land \perceives\Big(a, t, \lnot \holds\big(\vicinity(b, e),t\big)\Big) \\
& \hspace{90pt} \rightarrow \\
& \hspace{30pt} \believes\Big(a, t, \says\big(\gamma, b, t +\Delta,
\happens(e, t)\big)\Big)\\
\end{aligned}
\end{equation*}
\end{mdframed}
\end{small}

\noindent{\textbf{Proof Sketch}}: Colors used for readability. 

\begin{minipage}[t]{0.45\linewidth}
\begin{footnotesize} 
    \begin{prooftree}
      \AxiomC{$\perceives(a, t, \happens(e, t))$}
      \LeftLabel{$D_{[\perceives \leadsto \knows]}$}   \UnaryInfC{$\knows(a, t, \happens(e, t))$}        
     \LeftLabel{$D_{[\knows \leadsto \believes]}$}     \UnaryInfC{$\believes(a, t, \happens(e, t)) \equiv \boxed{\phi_1}$}   
  \end{prooftree}
\end{footnotesize}
\end{minipage}
\begin{minipage}[t]{0.45\linewidth}
\begin{footnotesize} 
    \begin{prooftree}
      \AxiomC{$\perceives(a, t, \lnot \holds\big(\vicinity(b, e),t\big))$}
   \LeftLabel{$D_{[\perceives \leadsto \knows]}$}   \UnaryInfC{$\knows\Big(a, t, \lnot \holds\big(\vicinity(b, e),t\big)\Big)$}        
     \LeftLabel{$D_{[\knows \leadsto \believes]}$}    \UnaryInfC{$\believes\Big(a, t, \lnot \holds\big(\vicinity(b, e),t\big)\Big) \equiv \boxed{\phi_2}$}   
  \end{prooftree}
\end{footnotesize}
\end{minipage}

Using $\mathbf{I}^f_3(ii)$ the CAIS observes all events that happen in
its enclosure:
\begin{footnotesize} 
\begin{prooftree}
        \AxiomC{$\mathbf{I}^f_3 (ii) \equiv\common\Bigg(\forall t, e: \Big[\happens(e, t) \leftrightarrow \perceives\big(\gamma, t, \happens(e, t)\big) \Big]\Bigg)$}
    \LeftLabel{$D_{[\common \leadsto \believes]}$  }
    \UnaryInfC{$\believes\left(a, t, \forall t, e: \Big[\happens(e, t)
            \leftrightarrow \perceives\big(\gamma, t, \happens(f, t)\big)
            \Big]\right)$}    \AxiomC{$\boxed{\phi_1}$}    
     \LeftLabel{$I_{\mathbf{B}}$}   \BinaryInfC{$\believes\Big(a, t, \perceives\big(\gamma, t,
          \happens(e, t)\big)\Big)$}   
      \LeftLabel{$I_{\mathbf{B}}$}  \UnaryInfC{$\believes \color{black}\big(\color{black} a, t, \color{violet}\believes\big(\gamma, t, \happens(e,t)\big) \color{black}\big)\color{black} \equiv
   \boxed{\psi_1}$}   
\end{prooftree}
\end{footnotesize} 
Similarly, using $\mathbf{I}^f_3 (iii)$:
\begin{footnotesize} 
\begin{prooftree}
        \AxiomC{$\mathbf{I}^f_3 (iii) \equiv \common\Bigg(\forall t,e : \Big[\lnot
          \holds(e, t)\leftrightarrow 
          \perceives\big(\gamma, t,  \lnot
          \holds(e, t)\big) \Big]\Bigg)$}
    \LeftLabel{$D_{[\common \leadsto \believes]}$  }  
\UnaryInfC{$\believes\left(a, t,\forall t, e\Big[\lnot
          \holds(e,t)\leftrightarrow 
          \perceives\big(\gamma, t,  \lnot
          \holds(e, t)\big)\Big]\right)$}        
\AxiomC{$\boxed{\phi_2}$} 
     \LeftLabel{$I_{\mathbf{B}}$}   \BinaryInfC{$\believes(a, t,
       \perceives(\gamma, t, \lnot 
          \holds\big(\vicinity(b, e), t)))$}   
    \LeftLabel{$I_{\mathbf{B}}$}     \UnaryInfC{$\believes(a, t,
      \believes\big(\gamma, t,  \lnot \holds\big(\vicinity(b, e), t\big))
      \equiv \boxed{\phi_3}$}   
\end{prooftree}
\end{footnotesize} 
From $\mathbf{I}^f_1$ (and from $\believes\big(a, t, \believes(\gamma, t, b \not = \gamma))$):
\begin{footnotesize} 
\begin{prooftree}
        \AxiomC{$\common\Big(\forall x, t,e: (x\not = \gamma)\rightarrow\big[
       \perceives\big(x, t,  \happens(e, t)\big) \rightarrow    \holds\big(\vicinity(x, e), t\big)
         \big]\Big)$}
    \LeftLabel{    $D_{[\common \leadsto \believes]}$  }
 \UnaryInfC{$\believes\big(a, t, \believes\big(\gamma, t, \forall x, t,e:  (x\not = \gamma)\rightarrow \big[
       \perceives\big(x, t,  \happens(e, t)\big) \rightarrow    \holds\big(\vicinity(x, e), t\big)
         \big]\big)\big)$}
\LeftLabel{    $I_{\believes}$  }
\UnaryInfC{$\believes\left(a, t, \believes\left(\gamma, t\Big[\lnot
          \holds\big(\vicinity(b, e), t\big)\rightarrow \lnot
          \perceives\big(b, t,  \happens(e, t)\big)\Big]\right)\right)$}        
\AxiomC{$\boxed{\phi_3}$} 
     \LeftLabel{$I_{\mathbf{B}}$}   \BinaryInfC{$\believes\Big(a, t,
       \believes\Big(\gamma, t, \lnot \perceives(b, t,
           \happens(e, t)\big)\Big)\Big)$}   
    \LeftLabel{$I_{\mathbf{B}}$}    
 \UnaryInfC{$\believes \color{black}\big(\color{black} a, t, \color{NavyBlue} \believes(\gamma, t, \lnot \believes(b,
  t, \happens(e,t))) \color{black}\big)\color{black}  \equiv
   \boxed{\psi_2}$}   
\end{prooftree}
\end{footnotesize} 
From $\mathbf{C}^f_2$, we have:
\begin{footnotesize} 
  \begin{prooftree}
    \AxiomC{   
      $ \common\left(t,\left[  
          \color{violet} \believes(\gamma, t, \happens(e,t)) \color{black} \land \color{NavyBlue}\believes(\gamma, t,
                \lnot \believes\big(b, t, \happens(e,t))) \color{black} \rightarrow 
                 \color{OliveGreen} \says(\gamma, b, t + \Delta, \happens(e,t))\color{black}
              \right]\right)$}
\LeftLabel{$D_{[\common \leadsto \believes]}$}\UnaryInfC{   
      $ \believes \color{black}\big(\color{black} a, t,\left[  
         \color{violet} \believes(\gamma, t, \happens(e,t)) \color{black} \land \color{NavyBlue}\believes(\gamma, t,
                \lnot \believes\big(b, t, \happens(e,t))) \color{black}\rightarrow 
               \color{OliveGreen} \says(\gamma, b, t + \Delta, \happens(e,t))\color{black}
              \right]\color{black}\big) \color{black} \equiv \boxed{\psi}$}
\end{prooftree}
\end{footnotesize} 
Using the above derived, $\boxed{\psi}$ and using $I_{\mathbf{B}}$:

\begin{footnotesize} 
  \begin{prooftree}

\AxiomC{$\boxed{\psi}$}
 \AxiomC{ $\believes \color{black}\big(\color{black} a, t, \color{violet}\believes\big(\gamma, t, \happens(e,t)\big) \color{black}\big)\color{black} \equiv
   \boxed{\psi_1}$}
\AxiomC{$\believes \color{black}\big(\color{black} a, t, \color{NavyBlue} \believes(\gamma, t, \lnot \believes(b,
  t, \happens(e,t))) \color{black}\big)\color{black}  \equiv
   \boxed{\psi_2}$}
 \LeftLabel{$I_{\mathbf{B}}$} \TrinaryInfC{$\believes(a, t,
   \color{OliveGreen} \says(\gamma, b, t + \Delta,
   \happens(e,t))\color{black})$}  

\end{prooftree}
\end{footnotesize} 
 $\blacksquare$


%% file: cognitive_blockworld_framework.tex
\section{Cognitive Polysolid Framework}

We now introduce the \emph{cognitive polysolid famework} (CPF), a
class of problems that we use for experiments.  From the framework, we
can generate specific \emph{cognitive polysolid world instantiations}
in which we then declare the number of blocks/solids, their
properties, and how these blocks/solids can be moved, as well as any
agents and their possible beliefs or knowledge.

The CPF subsumes the familiar ``blocks world,'' described for instance
in \cite{nilsson1980}, and long used for reasoning and planning in
purely extensional ways.  Briefly, CPF gives us a physical and
cognitive domain unlike the purely physical blocks world domain.  (The
formal logic used in \cite{nilsson1980} is purely extensional, as it's
simply first-order logic.)  Since the physical complexities of blocks
world problems have been well explored \citep{Gupta1991}, and since
this microworld has been has been used for benchmarking
\citep{SLANEY2001119}, we emphasize cognitive extensions of it.

A cognitive polysolid world instantiation contains some finite number
of blocks and a table large enough to hold all of them.  Each block is
\texttt{on} one other object; that object can be another block or the
table.  A block is said to be \texttt{clear} if there is no block that
is on top of it.  To move the blocks, an agent can either
\texttt{stack} (placing a block on the table on top of another block)
or \texttt{unstack} (taking a block that is on top of another block
and placing it on the table).  Before stacking the blocks, both need
to be clear; when unstacking, the top block must be clear beforehand.
After stacking the blocks, the bottom block is then not clear, and
after unstacking, it is then clear. Translating this description to
the \DCEC, we add two additional sorts and a constant, as well as some
new functions; our augmentation is shown below:

\vspace{-0.03in}
\begin{footnotesize}
  \begin{equation*}
    \begin{aligned}\
      &\Surface \sqsubset \Object \\
      &\Block \sqsubset \Surface \\
      &\ctable: \Surface \\
      &\on: \Block \times \Surface \rightarrow \Fluent
             \end{aligned} \hspace{30pt}
 \begin{aligned}
      &\clear: \Block \rightarrow \Fluent \\
      &\goal: \Boolean \times \Number \rightarrow \Boolean \\
      &\stack: \Block \times \Block \rightarrow \ActionType \\
      &\unstack: \Block \times \Block \rightarrow \ActionType
    \end{aligned}
  \end{equation*}
\end{footnotesize}
\vspace{-0.05in}

%% file: scenario.tex
\section{A Cognitive Polysolid World Simulation}

We start with a very elementary cognitive polysolid world (though this
work scales fine to larger numbers).  We have three identical blocks,
named $\ablock$, $\bblock$, and $\cblock$, which all start on the
table.  This is represented in the $\DCEC$ as:

\begin{footnotesize}
  \begin{center}
    \begin{tabular}{ c c }
      \holds(\on(\ablock, \ctable), 0) & \holds(\clear(\ablock), 0)\\
      \holds(\on(\bblock, \ctable), 0) & \holds(\clear(\bblock), 0)\\
      \holds(\on(\cblock, \ctable), 0) & \holds(\clear(\cblock), 0)
    \end{tabular}
  \end{center}
\end{footnotesize}

There are only two human agents, $\humana$ and $\humanb$, who have
knowledge about how the cognitive polysolid world works.  Using this
instantiation, we give the room two tasks to demonstrate its theory of
mind as required by the constraints specified above.  For both tasks,
we will use the following sequence of events to configure the world
for the two tasks:

\begin{footnotesize}
  \begin{enumerate}
  \item{\humana\ and \humanb\ enter the room}
  \item{\humana\ moves block \ablock\ onto block \bblock}
  \item{\humanb\ adds the goal of block \cblock\ on block \bblock}
  \item{\humanb\ leaves the room}
  \item{\humana\ moves block \ablock\ to the table}
  \item{\humana\ removes the goal for block \cblock\ and adds the goal of block \ablock\ on block \cblock}
  \item{\humana\ moves block \ablock\ onto block \cblock}
  \item{\humanb\ returns to the room}
  \item{\humanb\ tries to move \ablock\ to the table.}
  \end{enumerate}
\end{footnotesize}
For this simulation, all events and fluents inside the room are
considered to be in the vicinity of agents within the room, and none
of the events and fluents within the room are considered to be in the
vicinity of agents outside the room when they happen or hold.  

\input{modified_false_belief_task}

%% file: modified_false_belief_task.tex
\begin{figure}

   \begin{center}
     \begin{minipage}[b]{0.4\textwidth}
     \begin{flushleft}   
      \begin{footnotesize} \underline{\humana's goals and beliefs}\\
        \vspace{4pt}
       \textbf{goals}: $On(A,C)$\\
        \vspace{-4pt}
       \textbf{belief}:
     \end{footnotesize}
     \end{flushleft}    
       \begin{tikzpicture}[auto centering, background rectangle/.style={fill=white}, show background rectangle]
         \node[style=block] (C) {$C$};
         \node[style=empty] (E) [above=of C] {};
         \node[style=block] (B) [left=of C, left=.5cm of C] {$B$};
         \node[style=block] (A) [left=of B, left =.5 cm of B] {$A$};
       \end{tikzpicture}
     \end{minipage}
     \begin{minipage}[b]{0.4\textwidth}
     \begin{flushleft} 
      \begin{footnotesize} \underline{\humanb's  goals and beliefs}\\                          
        \vspace{4pt}
       \textbf{goals:} $On(C,B)$\\
        \vspace{-4pt}
       \textbf{belief}:
     \end{footnotesize}
     \end{flushleft}   
       \begin{tikzpicture}[auto centering, background rectangle/.style={fill=gray!25}, show background rectangle]
         \node[style=block] (B) {$B$};
         \node[style=block] (A) [above=of B] {$A$};
         \node[style=block] (C) [right=of B, right=.5cm of B] {$C$};
       \end{tikzpicture}
     \end{minipage}
   \end{center}
\caption{Visual representation of mental states of the agents
  (\humanb's state is grayed as he is not in the room).}
\label{fig:mom-example}
\end{figure}
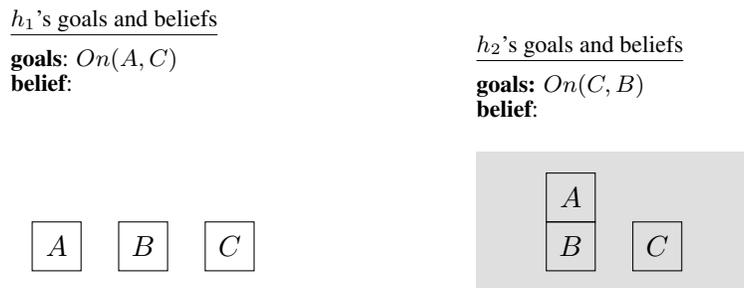
 
For the first portion of this task, we consider the world between
steps 5 \& 6.  At this point, we wish to see where the room believes
the blocks are, as well as where it believes that \humana\ and
\humanb\ think the blocks are, focusing primarily on block A.  We ask
the machine three questions, translating them into the \DCEC:
\emph{``Where does the CAIS/$\humana$/$\humanb$ believe block
  $\ablock$ is?''}.  We translate this question into three sentences
in the \DCEC\ which we can then pass down to \textsf{ShadowProver} to
answer.
For the first two questions, both the AI and the agent $\humana$ are
in the room and can perceive where the block is, and thus have
knowledge of its location.  $\humanb$ left the room at step 4 and
missed the block being moved at step 5.  Therefore, his knowledge of
where the block is remains at what it was when he was in the room. We
show the three statements below generated from \textsf{ShadowProver}
that answers the above questions as well as show a visual
representation of this answer in Figure \ref{fig:mom-example}:

\vspace{-0.03in}
\begin{footnotesize}
\begin{equation*}
\begin{aligned}\
&\believes(\cir, t, \holds(\on(\ablock, \ctable), t)) \\
&\believes(\humana, t, \holds(\on(\ablock, \ctable), t)) \\
&\believes(\humanb, t, \holds(\on(\ablock, \bblock), t))
\end{aligned}
\end{equation*}
\end{footnotesize}
\vspace{-0.05in}

For the second part of the task, we consider how the room responds
after step 9 (say at $t_{10}$).  The room believes that agent
\humanb\ still believes that the goal is $On(C, B)$, and that it
differs from the current goal of $On(A, C)$.  The room then notifies
\humanb\, showing a side-by-side comparison of the current goals of
the room (and the plan to achieve it) and what his believed goals with
respect to the room are (and the plan to achieve them).  The system is
able to do this, allowing for lag due to network calls, on average in
around 748ms employing \textsf{ShadowProver} and \textsf{Spectra}
multiple times.\footnote{The algorithmic details of the implementation
  are irrelevant and can be found in prior cited work.  There is a
  deeper integration with the reasoner and planner that allows us to
  offload intensive reasoning tasks, necessitated by the principles
  given above, as planning tasks.  Discussing this is beyond the scope
  of this paper.  A video of this task can be viewed at
  http://mpeveler.com/cbwf-falsebelief.html}
Note that this scenario is an instantation of \textbf{Property 1}
(Expectation of Usefulness) proved above. \textbf{Property 1} is
instantiated with the event $e$ being setting a new goal:
$e \equiv setGoal (On(A,C))$:

\begin{footnotesize}
\begin{equation*}
\begin{aligned}
&\perceives\big(\humana, t_6, \happens\big(setGoal(On(A,C)), t_6\big)\big) \land \perceives\Big(\humana, t_6, \lnot \holds\big(\vicinity(\humanb, setGoal (On(A,C))), t_6\big)\Big) \\
& \hspace{90pt} \rightarrow \believes\Big(\humana, t_9, \says\big(\gamma, \humanb, t_{10}
\happens(setGoal (On(A,C)), t_6)\big)\Big)
\end{aligned}
\end{equation*}
\end{footnotesize}

\section{A Quick Summary}
We give a quick summary before we delve into comparisons with
prior/related work below.

\begin{enumerate}[(1)]
\itemsep-0.80pt

\item In Section \ref{sect:prop_not_enough} we explained why systems that
  are less expressive than quantified modal logic cannot model beliefs
  and other mental states with fidelity.

\item We proposed in Section~\ref{sect:reqs} a pair of informal
  requirements $\langle\mathcal{C}, \mathcal{I} \rangle$ that a CAIS
  should satisfy.

\item In Section~\ref{sect:freqs}, we formalized these requirements in
  \DCEC, a quantified modal logic, resulting in:
  $$\langle \mathbf{C}^f_1, \mathbf{C}^f_2, \mathbf{I}^f_1,
  \mathbf{I}^f_2, \mathbf{I}^f_3\rangle$$

\item We then proved in Section~\ref{sect:prop} that these formal
  requirements lead to a desirable property: \textbf{Property 1} (a
  foundational property that can be used to build other such
  properties in future work).

\item Finally, we briefly presented an implementation of a system that
  adheres to the formalization.  We showed a scenario in a cognitive
  polysolid world where \textbf{Property 1} is useful and realized.
\end{enumerate}

%% file: objections.tex
\section{Prior/Related Work and Novelty}
\label{sect:prior_related_novelty}

\subsection{Microworlds}
\label{subsect:microworlds}

The environment for our case studies is, as we have said, \emph{not} a
blocks world.  Blocks worlds are all deficient from the point of view
of our research program, in significant part because the objects
within them are devoid of propositional attitudes, and hence mere
extensional logic suffices.  A classic confirmatory presentation and
lucid discussion of a classic blocks world can be found in
\cite{logical.foundations.ai}; it will be seen in that discussion that
the only objects in the microworld are inanimate and non-cognitive.
The same will be seen in early, seminal attempts to formalize physical
objects and processes, as for instance in
\cite{hayes_naive_physics_manifesto,hayes_2nd_naive_physics_man}.  In
stark contrast, CPF is a member of a class of microworlds best
lableled as \textit{cognitive microworlds}; a \textit{sine qua non}
for a microworld being in this class is that some of the objects
therein are cognitive agents that as such have propositional attitudes
and obligations, and attend, perceive, communicate at the level of
human natural language, etc.  Therefore a CPF includes instances in
which the environment is populated by agents the representation of
which requires \emph{intensional} logic, and indeed quantified
intensional logic.\footnote{In further contrast, CPF can include
  irreducibly visual entities the representation of which requires
  heterogeneous logic \cite{heterogeneous.logic,vivid_aij}, which none
  of the prior/related researchers in the logicist tradition referred
  to in the present section ever investigated, since they worked/work
  on straight symbolic systems, with no homomorphic representations to
  be found.  Moreover, even in the case of non-mental objects,
  arbitrary polysolids are allowed, and these solids can be in motion
  through time.  The case studies discussed in the present paper don't
  employ the full available range of entities allowable in a CPF; and
  formal specification of CPF and the broader category of cognitive
  microworlds is out of scope here.}

\subsection{Unprecedented Expressivity}
\label{subsect:unpreceented_expressivity}

We have previously proved [e.g.\ in
  \citep{selmer_naveen_metaphil_web_intelligence}] that we cannot
model even everyday propositional knowledge in non-intensional
systems.\footnote{The founders of modern AI, many of whom were
  logicits, all came to the field from mathematical logic, which is by
  definition extensional, not intensional; see
  e.g.\ \citep{ebb.flum.thomas.2nded} for discussion.  The irony here
  is that Leibniz can be viewed as the primogenitor of logicist AI,
  and he invented \emph{both} modern extensional and intensional
  formal logic.}
For the sort of problems we are interested in, minimally, first-order
logic, married to multiple intensional operators, is required.  For a
simple instance of this point, consider that we need to have
uncompromising representations of statements such as:

\begin{small}
\begin{enumerate}
  \item \emph{``There is no one in the room who believes that no one is in the room.''}
  \item \emph{``The organizer believed that the number of people in
    the room (7) was more than what was allowed, and hence had to ask
    some of the participants to leave.'' }
\end{enumerate}
\end{small}

It is impossible to model these statements in systems that are not at
once non-quantificational or non-intensional in nature.  Even prior
art such as the event calculus and the systems in
\cite{wooldridge.mas}, which are sensitive to expressivity demands,
are based on systems that are markedly less expressive than the
quantified multi-operator modal calculus we use in the present
paper.\footnote{For a more detailed discussion, please see the
  appendix in \cite{nsg_sb_dde_ijcai}.}  Along the same line, from the
formal point of view, the excellent, aforementioned
\cite{pearce_etal_social_planning_aaai2014} allows nested belief
operators, but their scope does not include unrestricted
quantification, rather sub-formulae in zero-order logic; and the other
intensional operators, epistemic and otherwise, are absent.  The
false-belief task was provably solved by the logicist AI system
specified and implemente3d in \cite{ka_sb_scc_seqcalc}, and even
infinitary cognitive challenges are met in \cite{epistemic.climav},
but this work is essentially a proper subset and ancestor of the
richer formalism and technology brought to bear in our attack on
CAISs.

\subsection{Rejection of Fixed Logics}
\label{subsect:rejection_fixed_logics}

Our work is not based on a particular, standard logic, such as are
brought to bear in the epistemic case in the likes of
\citep{moore_autoepistemiclogic_aij} and
\citep{reasoning.about.knowledge}, or in work that directly
appropriates a logic-programming base and syntax, since such bases are
invariably built atop extensional resolution.  Indeed, it's in part
precisely because of the inadequacy of ``off the shelf'' logics [such
  as those in the well-known $\Diamond$-$\Box$ ontology for
  modal/epistemic logics] that motivated the creation of cognitive
calculi in the first place.  As seen above, cognitive calculi are a
space of formal systems that are composed of a typed signature
$\Sigma$ (which in turn include an alphabet, formal grammar, and type
information), along with a tailor-made, easily adjustable collection
of inference schemata.  Hence, given cognitive calculus is unlikely to
correspond to any fixed, rigid logic; moreoever, any use of a informal
syntax renders the work based upon this use radically different than
our specification of signatures.  Almost all prior work in AI in the
logicist tradition is based on off-the-shelf logics, with standard,
long-standing inference rules.  In fact, the early, seminal work done
in this tradition by Newell, Simon, McCarthy, and Hayes was based on
exploitation of the propositional and predicate calculi.  Our approach
is dramatically different, in that \textit{any} collection of natural
(where `natural' is a here a technical term, one used in
e.g.\ `natural deduction) inference schemata can be created and used,
and immediately implemented via corresponding adjustments in
ShadowProver.  Note as well that inference schemata can be, and often
in our work are, meta-logical.\footnote{E.g.,~a schema might say that
  if $\phi$ is provable in less than $k$ steps, then an agent should
  believe $\phi$.} This enables what appears to be unprecedented
flexibility.

%

\subsection{Rejection of Standard Semantics/Models in Favor of
Proof-Theoretic Semantics}
\label{subsect:rejection_semantics}

Here is a telling quote from early work by Hayes:

  \begin{small}
    \begin{quote}
      The ability to interpret our axioms in a possible world, see
      what they say and whether it is true or not, is so useful that I
      cannot imagine proceeding without it.  $\ldots$ The main
      attraction of formal logics as representational languages is
      that they have very precise model theories, and the main
      attraction of first-order logic is that its model theory is so
      simple, so widely applicable, and yet so powerful.
      (Hayes 1985, 10)
   \end{quote}
  \end{small}

\noindent
This quote is telling because our cognitive calculi are purely
inference-theoretic, and are traceable in this regard directly back to
the advent of proof-theoretic semantics
\cite{prawitz_philosophical_position_proof_theory}, which eschews
semantics of the sort that Hayes venerates.  A cognitive calculus has
no provided model theory for its extensional levels, and rejects any
use for instance of possible worlds to provide non-inferential
semantics for modal operators, a rejection that can be traced back to
an early proof by \cite{are_there_set-theoretic_worlds} that standard
set-theoretic unpackings of possible worlds lead to absurdity.

\subsection{What About Bayesian Approaches?}
\label{subsect:bayesian_approaches}

We have of course placed the underlying technical content of our AI
work on CAISs within the logicist tradition.\footnote{We have for
  economy herein gone ``light'' on the historical trajectory of this
  approach (which is easily traceable to Leibniz, and even to
  Aristotle) to both modeling and mechanizing human-level cognition.
  For an overview of \emph{all} of AI, including from an historical
  point of view, and covering Bayesianism in the field as well, see
  \cite{sep_ai}.  Dedicated coverage of the logicist approach in AI
  and computational cognitive science can be found here:
  \citep{logicist_manifesto,rascals_in_sun}.}  What about Bayesian
approaches to reaching the goals we have for smart rooms and in
particular CAISs?  How does such work relate technically to cognitive
calculi and the reasoning and planning systems that work symbiotically
with them (all covered, of course, above)?  We do not have space for a
sustained answer to these questions, and therefore opt for stark
brevity via but two points: One, since Bayesian work in AI is of
necessity based on underlying formal languages of the extensional and
simple sort (e.g., propositional and predicate calculi with the key
function symbol \textit{prob} whose range is $[0,1]$) used to specify
at least parts of the probability calculus going back to Kolmogorov,
and since these formal languages are painfully inexpressive relative
to cognitive calculi, Bayesian formalisms and technology built atop
them are inadequate for reaching the goals we have set.\footnote{Even
  inductive logicians of the first rank readily acknowledge that the
  probability of such propositions as that Jones believes that Smith
  believes that Jones believes that there are exactly two properties
  shared between Black and Smith are inexpressible in Bayesianism;
  e.g.\ see \citep{fitelson_on_pollock_thinking_acting}.}  Two:
Bringsjord and Govindarajulu subsume probability into a
proof-theoretic approach based on multi-valued cognitive calculi,
where the values are likelihood values, or --- as they are sometimes
known --- strength factors.  For details, see
\citep{uncertaintyized_cognitive_calculus}.

\subsection{Novelty?}
\label{subsect:novelty}

While intelligent and immersive rooms have been designed and built by
researchers for decades, there has not been any formal, rigorous work
characterizing an intelligent room and how it differs from other
AI-infused environments.  We offer the first such characterization of
an intelligent room that has cognitive abilities: a CAIS powered by AI
operating on the strength of cognitive calculi, which are in turn
empowered by implemented reasoning and planning technology, where
specific requirements must be met.  We have also shown that a room
that satisfies these requirements can have a useful property that we
term \emph{expectation of usefulness}.  One advantage of our
formalization is that we have used a quantified multi-operator modal
logic that has been previously used to model higher-level cognitive
principles.  This opens up possibilities for principled integration of
other cognitive abilities in the future.

%% file: conclusion.tex
\section{Conclusion}

We now quickly summarize our contributions and present promising future lines of work. Our primary contribution is in the creation of an overseeing AI that is capable of tracking participants' mental states as they operate within a CAIS. This AI is capable of using this information to reason and plan assisting the participants in completing a task and tracking information as well as generate explanations for its actions. As part of this system, we give a definition for a concrete framework that can be used to generate future tests of increasing complexity for the system. We give the necessary machinery via syntax and sorts to use this framework as well as show a base implementation that was derived from the framework. Within this implementation, we show two tasks that can be solved via the the overarching AI that encompasses a CAIS and has a sufficient definition of theory of mind. 

	Future work will be on further empowering the overarching AI,  in creating additional domains of work, and to create an overarching formal definition for a \emph{cognitive and immersive framework}. To extend the AI framework, we are considering adding partial satisfaction planning capabilities~\citep{van2004effective,smith2004choosing} to the Spectra planner so that it can reason about goals that conflict with each other in a utility-centric framework. In addition, we currently allow agents to add goal priority explicitly. However, in a real group discussion, priority of a given goal may shift implicitly or by recognizing the particular belief states of the agents about a given goal. To improve upon this, we aim to incorporate research of social choice theory and ranking data via such methods as discussed in \cite{xia2017}.

	The \emph{cognitive and immersive framework} would form the basis for any of our cognitive frameworks, which includes the CPF. From the core cognitive definition of the CPF, we hope to expand our system to other domains that would benefit from the cognitive capabilities demonstrated here. For one such domain, the presented work can be very useful in business negotiation scenarios like contracts management or using software where multiple parties are involved and can have different vantage points (mental models) for discussion. There has been some work on analyzing contracts for identifying gaps~\citep{contracts-commit} and terms of conditions for software services~\citep{api-tac} but they do not consider parties' mental models. Finally, we aim to further enhance and refine the CPF such that it can be used to capture and reason about more complex domains than the presented blocks world, such as described in \cite{barker2017logical, johnson2016} and how agents may interact with them.

%% file: acknowledgments.tex
\section{Acknowledgments}

We are grateful for both IBM and Rensselaer support of the research
reported on herein, via both the CISL and AIRC programs.  Our
gratitude extends to both ONR and AFOSR, support from which has among
other things in part enabled the invention, implementation, and
refinement of \textsf{ShadowProver} and \textsf{Spectra}.  Four
anonymous referees provided helpful comments, and we deeply appreciate
their insights.  Trenchant, precise analysis provided by Pat Langley
proved to be invaluable, and we are indebted.


%% file: main.bbl
\begin{thebibliography}{52}
\expandafter\ifx\csname natexlab\endcsname\relax\def\natexlab#1{#1}\fi
\expandafter\ifx\csname url\endcsname\relax
  \def\url#1{{\path{\sloppy #1}}}\fi
\expandafter\ifx\csname urlprefix\endcsname\relax\def\urlprefix{From }\fi

\bibitem[{Arkoudas \& Bringsjord(2004)}]{epistemic.climav}
Arkoudas, K., \& Bringsjord, S. (2004).
\newblock {Metareasoning for multi-agent epistemic logics}.
\newblock {\em Proceedings of the Fifth International Conference on
  Computational Logic In Multi-Agent Systems ({CLIMA} 2004)\/} (pp. 50--65).
  Lisbon, Portugal.

\bibitem[{Arkoudas \& Bringsjord(2008)}]{ArkoudasAndBringsjord2008Pricai}
Arkoudas, K., \& Bringsjord, S. (2008).
\newblock {Toward formalizing common-sense psychology: {A}n analysis of the
  false-belief task}.
\newblock {\em {Proceedings of the Tenth Pacific Rim International Conference
  on Artificial Intelligence}\/} (pp. 17--29). Springer-Verlag.

\bibitem[{Arkoudas \& Bringsjord(2009{\natexlab{a}})}]{ka_sb_scc_seqcalc}
Arkoudas, K., \& Bringsjord, S. (2009{\natexlab{a}}).
\newblock {Propositional attitudes and causation}.
\newblock {\em International Journal of Software and Informatics\/}, {\em 3\/},
  47--65.

\bibitem[{Arkoudas \& Bringsjord(2009{\natexlab{b}})}]{vivid_aij}
Arkoudas, K., \& Bringsjord, S. (2009{\natexlab{b}}).
\newblock {Vivid: An AI framework for heterogeneous problem solving}.
\newblock {\em Artificial Intelligence\/}, {\em 173\/}, 1367--1405.

\bibitem[{Barker-Plummer et~al.(2017)Barker-Plummer, Barwise, \&
  Etchemendy}]{barker2017logical}
Barker-Plummer, D., Barwise, J., \& Etchemendy, J. (2017).
\newblock {\em Logical reasoning with diagrams and sentences: Using
  hyperproof\/}.
\newblock CSLI lecture notes. CSLI Publications/Center for the Study of
  Language \& Information.

\bibitem[{Barwise \& Etchemendy(1995)}]{heterogeneous.logic}
Barwise, J., \& Etchemendy, J. (1995).
\newblock Heterogeneous logic.
\newblock In J.~Glasgow, N.~Narayanan, \& B.~Chandrasekaran (Eds.), {\em
  Diagrammatic reasoning: {C}ognitive and computational perspectives\/},
  211--234. Cambridge, MA: MIT Press.

\bibitem[{Block(1995)}]{bbs.block}
Block, N. (1995).
\newblock {On a confusion about a function of consciousness}.
\newblock {\em Behavioral and Brain Sciences\/}, {\em 18\/}, 227--247.

\bibitem[{Bringsjord(1985)}]{are_there_set-theoretic_worlds}
Bringsjord, S. (1985).
\newblock {Are there set-theoretic worlds?}
\newblock {\em Analysis\/}, {\em 45\/}, 64.

\bibitem[{Bringsjord(2007)}]{sb_billion_conscious_robot}
Bringsjord, S. (2007).
\newblock {Offer: One billion dollars for a conscious robot. If you're honest,
  you must decline}.
\newblock {\em Journal of Consciousness Studies\/}, {\em 14\/}, 28--43.

\bibitem[{Bringsjord(2008{\natexlab{a}})}]{rascals_in_sun}
Bringsjord, S. (2008{\natexlab{a}}).
\newblock {Declarative/logic-based cognitive modeling}.
\newblock In R.~Sun (Ed.), {\em {The Handbook of Computational Psychology}\/},
  127--169. Cambridge, UK: Cambridge University Press.

\bibitem[{Bringsjord(2008{\natexlab{b}})}]{logicist_manifesto}
Bringsjord, S. (2008{\natexlab{b}}).
\newblock {The logicist manifesto: {A}t long last let logic-based {AI} become a
  field unto itself}.
\newblock {\em Journal of Applied Logic\/}, {\em 6\/}, 502--525.

\bibitem[{Bringsjord(2010)}]{sb_on_knowledge_game}
Bringsjord, S. (2010).
\newblock {Meeting Floridi's challenge to artificial intelligence from the
  knowledge-game test for self-consciousness}.
\newblock {\em Metaphilosophy\/}, {\em 41\/}, 292--312.

\bibitem[{Bringsjord et~al.(2018)Bringsjord, Bello, \&
  Govindarajulu}]{axiomatizing_consciousness1}
Bringsjord, S., Bello, P., \& Govindarajulu, N. (2018).
\newblock {Toward axiomatizing consciousness}.
\newblock In D.~Jacquette (Ed.), {\em The bloomsbury companion to the
  philosophy of consciousness\/},  289--324. London, UK: Bloomsbury Academic.

\bibitem[{Bringsjord \&
  Govindarajulu(2012)}]{selmer_naveen_metaphil_web_intelligence}
Bringsjord, S., \& Govindarajulu, N.~S. (2012).
\newblock {Given the web, what is intelligence, really?}
\newblock {\em Metaphilosophy\/}, {\em 43\/}, 361--532.

\bibitem[{Bringsjord \& Govindarajulu(2017)}]{sep_ai}
Bringsjord, S., \& Govindarajulu, N.~S. (2017).
\newblock {Artificial Intelligence}.
\newblock In E.~Zalta (Ed.), {\em {The Stanford Encyclopedia of Philosophy}\/}.
\newblock
  \urlprefix\url{https://plato.stanford.edu/entries/artificial-intelligence}.

\bibitem[{Bringsjord et~al.(2015)Bringsjord, Licato, Govindarajulu, Ghosh, \&
  Sen}]{roman2015_robot_self-con}
Bringsjord, S., Licato, J., Govindarajulu, N., Ghosh, R., \& Sen, A. (2015).
\newblock {Real robots that pass tests of self-consciousness}.
\newblock {\em Proccedings of the 24th IEEE International Symposium on Robot
  and Human Interactive Communication\/} (pp. 498--504). New York, NY: IEEE.

\bibitem[{Brooks(1997)}]{brooks1997intelligent}
Brooks, R.~A. (1997).
\newblock The intelligent room project.
\newblock {\em Proceedings of Second Internal Conference on Cognitive
  Technology, 1997. Humanizing the Information Age.\/} (pp. 271--278). IEEE,
  Aizu-Wakamatsu City, Japan: IEEE Computer Society Press.

\bibitem[{Chakraborti et~al.(2015)Chakraborti, Briggs, Talamadupula, Zhang,
  Scheutz, Smith, \& Kambhampati}]{chakraborti2015planning}
Chakraborti, T., Briggs, G., Talamadupula, K., Zhang, Y., Scheutz, M., Smith,
  D., \& Kambhampati, S. (2015).
\newblock Planning for serendipity.
\newblock {\em Proceedings of 2015 IEEE/RSJ International Conference on
  Intelligent Robots and Systems2015 IEEE/RSJ International Conference on
  Intelligent Robots and Systems\/} (pp. 5300--5306). IEEE, Hamburg, Germany:
  IEEE.

\bibitem[{Chakraborti et~al.(2017)Chakraborti, Talamadupula, Dholakia,
  Srivastava, Kephart, \& Bellamy}]{jones2017aaaifss}
Chakraborti, T., Talamadupula, K., Dholakia, M., Srivastava, B., Kephart,
  J.~O., \& Bellamy, R. K.~E. (2017).
\newblock {Mr. Jones -- Towards a proactive smart room orchestrator}.
\newblock {\em AAAI Fall Symposium on Human-Agent Groups\/}.

\bibitem[{Chan et~al.(2008)Chan, Est{\`e}ve, Escriba, \&
  Campo}]{chan2008review}
Chan, M., Est{\`e}ve, D., Escriba, C., \& Campo, E. (2008).
\newblock A review of smart homes—present state and future challenges.
\newblock {\em Computer methods and programs in biomedicine\/}, {\em 91\/},
  55--81.

\bibitem[{Cirillo et~al.(2009)Cirillo, Karlsson, \&
  Saffiotti}]{cirillo2009human}
Cirillo, M., Karlsson, L., \& Saffiotti, A. (2009).
\newblock Human-aware task planning for mobile robots.
\newblock {\em Proceedings of 2009 International Conference on Advanced
  Robotics\/} (pp. 1--7). IEEE.

\bibitem[{Coen et~al.(1998)}]{coen1998design}
Coen, M.~H., et~al. (1998).
\newblock {Design principles for intelligent environments}.
\newblock {\em Proceedings of the Fifteenth National/Tenth Conference on
  Artificial Intelligence/Innovative Applications of Artificial Intelligence\/}
  (pp. 547--554). Madison, Wisconsin, USA: AAAI Press.

\bibitem[{Desai et~al.(2008)Desai, Narendra, \& Singh}]{contracts-commit}
Desai, N., Narendra, N.~C., \& Singh, M.~P. (2008).
\newblock {Checking correctness of business contracts via commitments}.
\newblock {\em Proceedings of 7th Int. Conf. on Autonomous Agents and
  Multiagent Systems\/}.

\bibitem[{Divekar et~al.(2018)Divekar, Peveler, Rouhani, Zhao, Kephart, Allen,
  Wang, Ji, \& Su}]{divekar2018}
Divekar, R.~R., Peveler, M., Rouhani, R., Zhao, R., Kephart, J.~O., Allen, D.,
  Wang, K., Ji, Q., \& Su, H. (2018).
\newblock Cira---an architecture for building configurable immersive
  smart-rooms.
\newblock {\em In Proceedings of Intellisys 2018\/}.

\bibitem[{Ebbinghaus et~al.(1994)Ebbinghaus, Flum, \&
  Thomas}]{ebb.flum.thomas.2nded}
Ebbinghaus, H.~D., Flum, J., \& Thomas, W. (1994).
\newblock {\em Mathematical logic (second edition)\/}.
\newblock New York, NY: Springer-Verlag.

\bibitem[{Fagin et~al.(2004)Fagin, Halpern, Moses, \&
  Vardi}]{reasoning.about.knowledge}
Fagin, R., Halpern, J., Moses, Y., \& Vardi, M. (2004).
\newblock {\em Reasoning about knowledge\/}.
\newblock Cambridge, MA: MIT Press.

\bibitem[{Fitelson(2010)}]{fitelson_on_pollock_thinking_acting}
Fitelson, B. (2010).
\newblock {Pollock on Probability in Epistemology}.
\newblock {\em Philosophical Studies\/}, {\em 148\/}, 455--465.

\bibitem[{Frith \& Frith(2005)}]{frith2005}
Frith, C., \& Frith, U. (2005).
\newblock Theory of mind.
\newblock {\em Current Biology\/}, {\em 15\/}, R644--R645.

\bibitem[{Genesereth \& Nilsson(1987)}]{logical.foundations.ai}
Genesereth, M., \& Nilsson, N. (1987).
\newblock {\em Logical foundations of artificial intelligence\/}.
\newblock Los Altos, CA: Morgan Kaufmann.

\bibitem[{Gentzen(1964)}]{gentzen1964}
Gentzen, G. (1964).
\newblock Investigations into logical deduction.
\newblock {\em American Philosophical Quarterly\/}, {\em 1\/}, 288--306.

\bibitem[{Govindarajulu \& Bringsjord(2017{\natexlab{a}})}]{nsg_sb_dde_ijcai}
Govindarajulu, N., \& Bringsjord, S. (2017{\natexlab{a}}).
\newblock {On automating the doctrine of double effect}.
\newblock {\em Proceedings of the Twenty-Sixth International Joint Conference
  on Artificial Intelligence\/} (pp. 4722--4730). Melbourne, Australia:
  International Joint Conferences on Artificial Intelligence.

\bibitem[{Govindarajulu \&
  Bringsjord(2017{\natexlab{b}})}]{uncertaintyized_cognitive_calculus}
Govindarajulu, N.~S., \& Bringsjord, S. (2017{\natexlab{b}}).
\newblock {Strength factors: An uncertainty system for quantified modal logic}.
\newblock {\em {Proceedings of the IJCAI Workshop on ``Logical Foundations for
  Uncertainty and Machine Learning}\/} (pp. 34--40). Melbourne, Australia.

\bibitem[{Gupta \& Nau(1991)}]{Gupta1991}
Gupta, N., \& Nau, D.~S. (1991).
\newblock Complexity results for blocks-world planning.
\newblock {\em Proceedings of the Ninth National Conference on Artificial
  Intelligence - Volume 2\/} (pp. 629--633). Anaheim, California: AAAI Press.

\bibitem[{Hayes(1978)}]{hayes_naive_physics_manifesto}
Hayes, P. (1978).
\newblock {The Na{\"{i}}ve physics manifesto}.
\newblock In D.~Mitchie (Ed.), {\em Expert systems in the microeletronics
  age\/},  242--270. Edinburgh, Scotland: Edinburgh University Press.

\bibitem[{Hayes(1985)}]{hayes_2nd_naive_physics_man}
Hayes, P.~J. (1985).
\newblock {The Second na{\"{i}}ve physics manifesto}.
\newblock In J.~R. Hobbs \& B.~Moore (Eds.), {\em Formal theories of the
  commonsense world\/},  1--36. Norwood, NJ: Ablex.

\bibitem[{Johnson et~al.(2016)Johnson, Hariharan, van~der Maaten, Fei{-}Fei,
  Zitnick, \& Girshick}]{johnson2016}
Johnson, J., Hariharan, B., van~der Maaten, L., Fei{-}Fei, L., Zitnick, C.~L.,
  \& Girshick, R.~B. (2016).
\newblock {CLEVR:} {A} diagnostic dataset for compositional language and
  elementary visual reasoning.
\newblock {\em CoRR\/}, {\em abs/1612.06890\/}.
\newblock \urlprefix\url{http://arxiv.org/abs/1612.06890}.

\bibitem[{Kim \& Shah(2017)}]{kim2017aaaifss}
Kim, J., \& Shah, J. (2017).
\newblock {Towards intelligent decision support in human team planning}.
\newblock {\em AAAI Fall Symposium on Human-Agent Groups\/}.

\bibitem[{Langley et~al.(2016)Langley, Barley, Choi, Katz, \&
  Meadows}]{langley2016pug}
Langley, P., Barley, M., Choi, D., Katz, E., \& Meadows, B. (2016).
\newblock Goals, utilities, and mental simulation in continuous planning.
\newblock {\em Proceedings of the Fourth Annual Conference on Advances in
  Cognitive Systems\/}.

\bibitem[{Moore(1985)}]{moore_autoepistemiclogic_aij}
Moore, R. (1985).
\newblock {Semantical considerations on nonmonotonic logic}.
\newblock {\em Artificial Intelligence\/}, {\em 25\/}, 75--94.

\bibitem[{Mueller(2014)}]{mueller2014}
Mueller, E. (2014).
\newblock {\em Commonsense reasoning: An event calculus based approach\/}.
\newblock Morgan Kaufmann, second edition.

\bibitem[{Nilsson(1980)}]{nilsson1980}
Nilsson, N.~J. (1980).
\newblock {\em Principles of artificial intelligence\/}.
\newblock San Francisco, CA, USA: Morgan Kaufmann Publishers Inc.

\bibitem[{Pearce et~al.(2014)Pearce, Meadows, Langley, \&
  Barley}]{pearce_etal_social_planning_aaai2014}
Pearce, C., Meadows, B., Langley, P., \& Barley, M. (2014).
\newblock Social planning: Achieving goals by altering others’ mental states.
\newblock {\em Proceedings of the 28th {AAAI} Conference on Artificial
  Intelligence\/} (pp. 402--408). Palo Alto, CA: AAAI.

\bibitem[{Prawitz(1972)}]{prawitz_philosophical_position_proof_theory}
Prawitz, D. (1972).
\newblock {The philosophical position of proof theory}.
\newblock In R.~E. Olson \& A.~M. Paul (Eds.), {\em Contemporary philosophy in
  scandinavia\/},  123--134. Baltimore, MD: Johns Hopkins Press.

\bibitem[{Premack \& Woodruff(1978)}]{PremackWoodruff78}
Premack, D., \& Woodruff, G. (1978).
\newblock {Does the chimpanzee have a theory of mind?}
\newblock {\em Behavioral and Brain Sciences\/}, {\em 4\/}, 515--526.

\bibitem[{Sengupta et~al.(2017)Sengupta, Chakraborti, Sreedharan, \&
  Kambhampati}]{radar2017aaaifss}
Sengupta, S., Chakraborti, T., Sreedharan, S., \& Kambhampati, S. (2017).
\newblock {RADAR - A proactive decision support system for human-in-the-loop
  planning}.
\newblock {\em AAAI Fall Symposium on Human-Agent Groups\/}.

\bibitem[{Slaney \& Thiébaux(2001)}]{SLANEY2001119}
Slaney, J., \& Thiébaux, S. (2001).
\newblock Blocks world revisited.
\newblock {\em Artificial Intelligence\/}, {\em 125\/}, 119 -- 153.

\bibitem[{Smith(2004)}]{smith2004choosing}
Smith, D.~E. (2004).
\newblock Choosing objectives in over-subscription planning.
\newblock {\em Proceedings of the Fourteenth International Conference on
  International Conference on Automated Planning and Scheduling\/} (p. 393).

\bibitem[{Talamadupula et~al.(2014)Talamadupula, Briggs, Chakraborti, Scheutz,
  \& Kambhampati}]{talamadupula2014coordination}
Talamadupula, K., Briggs, G., Chakraborti, T., Scheutz, M., \& Kambhampati, S.
  (2014).
\newblock 2014 ieee/rsj international conference on intelligent robots and
  systems.
\newblock {\em 2014 IEEE/RSJ International Conference on Intelligent Robots and
  Systems\/} (pp. 2957--2962). Chicago, IL: IEEE.

\bibitem[{Van Den~Briel et~al.(2004)Van Den~Briel, Sanchez, Do, \&
  Kambhampati}]{van2004effective}
Van Den~Briel, M., Sanchez, R., Do, M.~B., \& Kambhampati, S. (2004).
\newblock Effective approaches for partial satisfaction (over-subscription)
  planning.
\newblock {\em Proceedings of the Nineteenth National Conference on Artificial
  Intelligence\/} (pp. 562--569). San Jose, CA: AAAI Press.

\bibitem[{Vukovic et~al.(2014)Vukovic, Laredo, \& Rajagopal}]{api-tac}
Vukovic, M., Laredo, J., \& Rajagopal, S. (2014).
\newblock {API terms and conditions as a service}.
\newblock {\em Proceedings of 2014 Services Computing Conference\/}.

\bibitem[{Wooldridge(2002)}]{wooldridge.mas}
Wooldridge, M. (2002).
\newblock {\em An introduction to multi agent systems\/}.
\newblock Cambridge MA: MIT Press.

\bibitem[{Xia(2017)}]{xia2017}
Xia, L. (2017).
\newblock Improving group decision-making by artificial intelligence.
\newblock {\em Proceedings of the Twenty-Sixth International Joint Conference
  on Artificial Intelligence\/} (pp. 5156--5160). Melbourne, Australia.

\end{thebibliography}
